\documentclass{article}

 \usepackage[dblblindworkshop, final]{neurips_2026}
\workshoptitle{AI Data Readiness for Scientific Discovery} 

\usepackage[utf8]{inputenc} 
\usepackage[T1]{fontenc}    
\usepackage{hyperref}       
\usepackage{url}            
\usepackage{amsfonts}       
\usepackage{nicefrac}       
\usepackage{microtype}      
\usepackage{xcolor}         
\usepackage{amsmath}
\usepackage{booktabs,tabularx,array}
\usepackage{tikz}
\usepackage{caption}
\usepackage{float}
\usetikzlibrary{patterns}

\title{Predicting Unseen Scientific Interactions: Conditional Flow Models Hallucinate Non-Additive Effects}

\author{%
Sam Gijsen\thanks{\begin{tabular}[t]{@{}l@{}}
  Correspondence to: \texttt{samgijsen@gmail.com} \\
  Code available at: \url{https://github.com/SamGijsen/interaction-flows}
\end{tabular}} \hspace{1cm}
  Moritz Seiler \hspace{1cm}
  Kerstin Ritter \\
  \\
  Hertie Institute for AI in Brain Health, University of T\"ubingen, Germany \\
  T\"ubingen AI Center, University of T\"ubingen,  Germany \\
  Charit\'e -- Universit\"atsmedizin Berlin, Department of Psychiatry and Psychotherapy, Germany \\
  German Center for Mental Health (DZPG), partner site T\"ubingen, Germany
}

\begin{document}

\maketitle

\begin{abstract}
Conditional generative models are increasingly used for scientific prediction beyond the direct training support. Yet scientific questions often concern whether one intervention changes the effect of another. Such interactions can be of primary interest while explaining little of the total variation, allowing models to appear accurate overall without recovering the relevant scientific effect.  It remains unclear how these models learn interactions and when these successfully extrapolate to unseen conditions. We therefore study interaction recovery in flow-matching models using a physiological time-series simulator, drug dose-response data, combinatorial cell perturbations, and molecular conformer generations, where we separate recovery from hallucinated interaction structure. In the controlled setting, individual effects are learned much earlier than interactions and models hallucinate interactions where none exist. Across settings, we find qualitatively different generalization problems, with successful prediction depending on whether condition structure continues from observed to held-out targets. When it does, interactions are recoverable. If not, models can generate plausible responses while missing the interaction. We furthermore find that hallucinations increase as targets become less identifiable, and that successful interaction recovery in output space can fail under nonlinear readouts. Finally, we show that aggregate fidelity metrics can conceal whether models learn and extrapolate interactions.

\end{abstract}

Conditional generative models such as diffusion and flow matching models have gained significant attention for generating novel biological samples and for predicting biological responses under unseen experimental conditions. Prominent use cases include single-cell transcriptomics \citep{he2026squidiff,yu2026scdfm}, neuroscience \citep{gijsen2026flow}, and protein design \citep{watson2023rfdiffusion,ingraham2023chroma}.

To be scientifically valuable, models must be able to predict how experimental conditions interact, rather than only generating broadly plausible responses. For example, administering two drugs together may be antagonistic, independent, or synergistic, and redundant pairs require predicting that a drug has a clear effect on its own but not when combined with another. Importantly, however, the bulk of overall variability across samples is often dominated by effects of individual factors. Conversely, interactions often account only for a small fraction, despite regularly being the primary object of scientific interest. For instance, in single-cell perturbation experiments, the interaction between cell-type and perturbation determines less of the variance than either factor by itself, yet it is often the central question \citep{schaefer2026principled}. 

As aggregate fidelity metrics (such as MSE or FID) are heavily influenced by those stronger effects of individual factors, they may tell us relatively little about how interactions are learned during training or whether an interaction learned from observed combinations can be extrapolated to unseen ones. This is critical for scientific prediction and discovery, as this implies risks where models may appear converged while still learning the interaction, or appear reasonably accurate while predicting the effect that motivated an experiment incorrectly. Furthermore, interactions are not only a property of the system but also of the data representation. A nonlinear readout of a sample can imply a different interaction, and therefore interaction recovery in output space does not transfer to a quantity which often can be more scientifically relevant.

The difficulty of this generalization problem directly depends on the available training data. A held-out target may be identifiable via smoothness of the condition space, via graph-relational knowledge, or underdetermined if only marginal support is provided. 
The latter problem, often called compositional generalisation \citep{wiedemer2023compositional}, is fundamentally different in the presence of interactions, as a training support covering each factor alone no longer determines the response in the held-out region of condition space \citep{brady2025interaction}. Simply put, observing two factors separately does not determine how they interact \citep{lachapelle2023additive, dong2022first}. In this case, the missing relationship must come from observed joint conditions, inductive biases in the model, or meta-knowledge \citep{shojaee2025llm}.

Here, we study how conditional flow models learn and extrapolate non-additive interactions in four systems that differ in the manner in which the training support constrains the held-out targets. Specifically, 1) a photoplethysmography (PPG) simulator with ground truth and tunable interaction strength, 2) drug dose-response screens where observed doses surround the held-out conditions,  3) combinatorial perturbations in single cells where components of a held-out pair are optionally excluded from observed pairs, and 4) radii of gyration over generated conformers given distinct molecular supports.

\textbf{We summarize our contributions as follows:}
\begin{itemize}
    \item \textbf{Diagnostics for interaction prediction.} We decompose predicted interactions into interaction recovery and orthogonal hallucination, with an identity to interaction error.
    \item \textbf{Hallucination of non-additivity.} In a simulator with a controlled interaction strength, we find flow models to generate interactions even in fully additive regimes. We additionally find stronger hallucinations for less constrained targets across settings. 
    \item \textbf{Continuation rather than distance determines extrapolation.} We show that held-out regions of a shared condition space at equal distance from the training support can be recovered almost completely or barely at all, depending on the continuation of the condition structure off the training support. 
    \item \textbf{Aggregate metrics conceal failures to predict interactions.} We observe distinct learning trajectories for interactions compared to main effects, that low endpoint errors can accompany failed interaction recovery, and that recovery in output space can fail after a scientific readout.
\end{itemize}

\section{Related work}

\textbf{Predicting perturbation and combination responses.}
Deep models are now commonly used to predict responses to unseen chemical and genetic perturbations and their combinations \citep{roohani2024predicting,klein2025cellflow,preuer2018deepsynergy}. However, recent work shows that such models can fail to beat additive or mean baselines for double perturbations \citep{ahlmann2025deep}, and that common metrics are inflated by systematic variation \citep{vinas2025systema}. As additive baselines predict no interaction, we use these as a reference for scoring interaction errors. We complement these existing evaluations by decomposing interaction errors and relating them to the identifiability of held-out condition regions.

\textbf{Compositional generalization and inductive biases of generative models.}
Compositional generalization has often been studied in settings where the joint response is determined by its components \citep{wiedemer2023compositional, okawa2023compositional, liu2022composable, lippl2024does}, which is an assumption that the presence of interactions violate. Prior work additionally links the order in which concepts are learned to their signal strength \citep{rahaman2019spectral,park2024emergence}, which invites the question how low-energy interactions are learned. Diffusion models have also been shown to produce unseen samples by interpolating between modes \citep{aithal2024understanding}, with the interaction hallucinations we study here constituting an analogous failure over condition space rather than sample space.

\textbf{Interactions depend on representations.}
Whether factors interact depends on the data representation in which we measure, which has been long established in epidemiology, where it is described as the interaction scale \citep{rothman1980concepts, vanderweele2014tutorial}. We extend this to the evaluation of generative models, showing that interaction recovery can differ between the output space and under a scientific readout.

\section{Methods}

\definecolor{coral}{HTML}{DC0073}
\definecolor{gridfill}{HTML}{FAF9F5}
\definecolor{gridline}{HTML}{C4C2B6}
\definecolor{gridedge}{HTML}{B4B2A9}
\definecolor{labelgray}{HTML}{5F5E5A}
\definecolor{inkdark}{HTML}{3A3A38}
 
\newcommand{\passmark}[2]{%
  \draw[inkdark,line width=1.0pt,line cap=round,line join=round]
    (#1,#2) -- ++(0.085,-0.095) -- ++(0.185,0.235);}
\newcommand{\failmark}[2]{%
  \draw[inkdark,line width=1.0pt,line cap=round] (#1,#2) -- ++(0.21,0.21);
  \draw[inkdark,line width=1.0pt,line cap=round] (#1,{#2+0.21}) -- ++(0.21,-0.21);}
 
\newcommand{\blobpath}{plot[smooth cycle,tension=0.5] coordinates
  {(3.992,1.320) (3.987,1.462) (3.964,1.603) (3.921,1.743) (3.852,1.879) (3.751,2.004)
   (3.615,2.112) (3.446,2.196) (3.253,2.250) (3.047,2.276) (2.843,2.277) (2.653,2.263)
   (2.483,2.248) (2.331,2.243) (2.191,2.255) (2.050,2.286) (1.897,2.332) (1.726,2.383)
   (1.534,2.427) (1.327,2.451) (1.115,2.448) (0.908,2.415) (0.717,2.351) (0.546,2.263)
   (0.398,2.156) (0.274,2.034) (0.172,1.902) (0.095,1.762) (0.046,1.617) (0.032,1.468)
   (0.057,1.320) (0.124,1.179) (0.228,1.050) (0.361,0.938) (0.510,0.842) (0.659,0.761)
   (0.799,0.687) (0.923,0.613) (1.033,0.533) (1.137,0.444) (1.245,0.349) (1.369,0.254)
   (1.514,0.170) (1.680,0.107) (1.862,0.072) (2.050,0.067) (2.236,0.090) (2.412,0.134)
   (2.577,0.189) (2.735,0.248) (2.890,0.307) (3.047,0.364) (3.209,0.423) (3.372,0.491)
   (3.530,0.571) (3.673,0.667) (3.793,0.779) (3.884,0.905) (3.945,1.039) (3.980,1.179)}}
\newcommand{\neckA}{(4.947,3.198) -- (-2.222,0.301)}
\newcommand{\neckB}{(3.303,-0.228) -- (3.081,2.937)}
\newcommand{\respcurve}[1]{%
  \draw[#1,line width=0.75pt,line join=round] plot[smooth,tension=0.6] coordinates
    {(0.10,0.14) (0.30,0.58) (0.46,0.70) (0.59,0.41) (0.73,0.51) (0.93,0.26)
     (1.18,0.13) (1.42,0.11) (1.62,0.23) (1.80,0.52) (1.94,0.66)};}

\begin{figure}[t]
\centering
\resizebox{0.85\textwidth}{!}{%
\begin{tikzpicture}[
  qlab/.style={circle,font=\tiny,text=white,inner sep=0pt,minimum size=7.2pt,
               fill=inkdark,draw=white,line width=0.6pt},
  qopen/.style={circle,font=\tiny,text=inkdark,inner sep=0pt,minimum size=7.2pt,
                fill=white,draw=inkdark,line width=0.7pt},
  flow/.style={draw=labelgray,line width=0.7pt,->}]
 
\node[anchor=base] at (2.23,3.92) {\small 1. Held-out queries};
\node[anchor=base] at (6.15,3.92) {\small 2. Aggregate check};
\node[anchor=base] at (10.50,3.92) {\small 3. Interaction geometry};
\node[anchor=base,text=labelgray] at (2.23,3.58) {\footnotesize condition space $\mathcal{C}$};
\node[anchor=base,text=labelgray] at (10.50,3.58) {\footnotesize $\hat{I}=\alpha I+I_{\perp}$};
 
\begin{scope}[shift={(0.15,0.45)}]
  \begin{scope}
    \clip \blobpath;
    \fill[gridfill] (-0.5,-0.5) rectangle (4.6,3.1);
    \fill[pattern=north east lines,pattern color=coral]
      (4.947,3.198) -- (-2.222,0.301) -- (-3.346,3.083) -- (3.823,5.979) -- cycle;
    \fill[pattern=north east lines,pattern color=coral]
      (3.303,-0.228) -- (3.081,2.937) -- (6.074,3.146) -- (6.295,-0.019) -- cycle;
    \draw[coral,line width=0.7pt] \neckA;
    \draw[coral,line width=0.7pt] \neckB;
  \end{scope}
  \draw[gridedge,line width=0.6pt] \blobpath;
 
  \foreach \bx/\by/\dx/\dy in {1.362/1.750/-0.157/0.389, 3.192/1.354/0.419/0.029}{
    \draw[white,line width=2.2pt] (\bx,\by) -- ++(\dx,\dy);
    \draw[inkdark,line width=0.6pt] (\bx,\by) -- ++(\dx,\dy);}
  \node[qlab] at (1.205,2.139) {1};
  \node[qlab] at (3.611,1.383) {2};
\end{scope}
 
\fill[gridfill] (0.55,-0.20) rectangle (0.71,-0.04);
\draw[gridedge,line width=0.4pt] (0.55,-0.20) rectangle (0.71,-0.04);
\node[anchor=base west,text=labelgray] at (0.79,-0.17) {\footnotesize observed};
\fill[pattern=north east lines,pattern color=coral] (2.30,-0.20) rectangle (2.46,-0.04);
\draw[coral,line width=0.5pt] (2.30,-0.20) rectangle (2.46,-0.04);
\node[anchor=base west,text=labelgray] at (2.54,-0.17) {\footnotesize held out};
 
\draw[flow] (4.42,1.88) -- (4.80,1.88);
 
\draw[gridline,line width=0.8pt] (4.85,3.62) -- (5.01,3.62);
\node[anchor=base west,text=labelgray,font=\scriptsize] at (5.05,3.58) {truth};
\draw[inkdark,line width=0.8pt] (6.15,3.62) -- (6.31,3.62);
\node[anchor=base west,text=labelgray,font=\scriptsize] at (6.35,3.58) {generated};
 
\node[qlab] at (4.98,2.775) {1};
\begin{scope}[shift={(5.15,2.30)}]
  \fill[gridfill,rounded corners=1.5pt] (0,0) rectangle (2.0,0.95);
  \draw[gridedge,line width=0.4pt,rounded corners=1.5pt] (0,0) rectangle (2.0,0.95);
  \begin{scope}[yshift=1.2pt]\respcurve{gridline}\end{scope}
  \begin{scope}[xshift=0.6pt,yshift=2.6pt]\respcurve{inkdark}\end{scope}
\end{scope}
\passmark{5.15}{2.01}
\node[anchor=base west,text=labelgray] at (5.46,1.96)
  {\footnotesize $\mathcal{E}_{\mathrm{end}}\ll 1$};
 
\node[qlab] at (4.98,1.325) {2};
\begin{scope}[shift={(5.15,0.85)}]
  \fill[gridfill,rounded corners=1.5pt] (0,0) rectangle (2.0,0.95);
  \draw[gridedge,line width=0.4pt,rounded corners=1.5pt] (0,0) rectangle (2.0,0.95);
  \begin{scope}[yshift=1.2pt]\respcurve{gridline}\end{scope}
  \begin{scope}[xshift=0.6pt,yshift=3.8pt]\respcurve{inkdark}\end{scope}
\end{scope}
\passmark{5.15}{0.56}
\node[anchor=base west,text=labelgray] at (5.46,0.51)
  {\footnotesize $\mathcal{E}_{\mathrm{end}}\ll 1$};
 
\draw[flow] (7.36,1.88) -- (7.70,1.88);
 
\begin{scope}[shift={(8.35,0.70)}]
  \draw[labelgray,line width=0.5pt] (-0.16,0) -- (4.34,0);
  \draw[labelgray,line width=0.5pt] (0,-0.16) -- (0,2.54);
  \draw[labelgray,line width=0.5pt] (2.0,-0.09) -- (2.0,0.09);
  \node[anchor=north,text=labelgray,font=\scriptsize] at (2.0,-0.11) {1};
  \draw[labelgray,line width=0.5pt] (-0.09,2.0) -- (0.09,2.0);
  \node[anchor=east,text=labelgray,font=\scriptsize] at (-0.11,2.0) {1};
  \draw[labelgray,line width=0.7pt,->,dash pattern=on 3.0pt off 1.6pt]
    (0.34,1.99) .. controls (1.55,2.32) and (2.85,1.90) .. (3.62,1.16);
  \node[anchor=south,text=labelgray] at (1.05,2.14) {\footnotesize $g$};
\end{scope}
 
\node[qlab]  at (8.51,2.64) {1};  \passmark{8.66}{2.38}
\node[qlab]  at (9.21,1.32) {2};
\node[anchor=south,text=inkdark] at (9.21,1.44) {\footnotesize $\approx$};
\node[qopen] at (12.15,1.70) {1};
\failmark{12.42}{1.60}
 
\node[anchor=base,text=labelgray] at (10.55,0.16)
  {\footnotesize hallucination $\beta$};
\node[rotate=90,anchor=base,text=labelgray] at (7.88,1.90)
  {\footnotesize recovery $\alpha$}; 
\end{tikzpicture}
} 
\caption{Aggregate generation quality does not guarantee accurate predicted interactions. Two conditions are sampled with equal distance off the training support but in different directions (1). Both samples score well on normalized endpoint error ($\mathcal{E}_\text{end}$), far below a trivial baseline (2). But decomposing the prediction into the component along the true interaction ($\alpha$) and the orthogonal, unrelated component ($\beta$) distinguishes the samples into interaction recovery ($\checkmark$) and attenuated interaction with hallucinated structure ($\approx$) (3). Finally, applying a readout function $g$ can considerably change the results ($\checkmark \xrightarrow{g}\times$), with less recovery and yet more hallucinated structure in the readout space compared to output space.}
\label{fig:concept}
\end{figure}
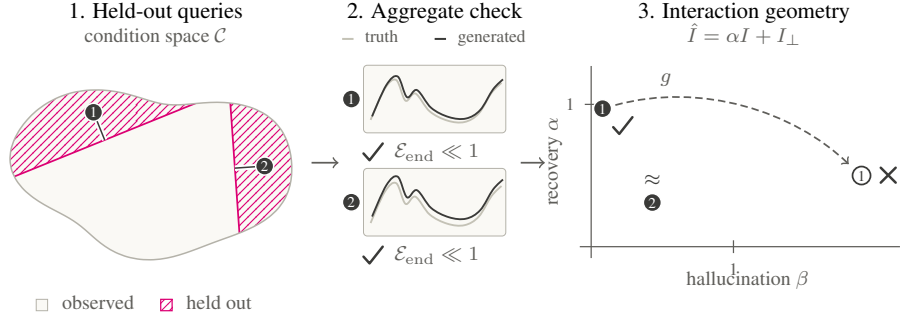

\subsection{Setup}
\label{sec:setup}

We start by providing a brief general framework. Given two factors $A$ and $B$
taking values $a$ and $b$ (categorical or continuous), flow matching models predict response
distributions $P_{a,b}$ over $\mathbb{R}^d$ with conditional means
$\mu(a,b) = \mathbb{E}[Y \mid a, b]$. The factors are additive if
$\mu(a,b) = g(a) + h(b)$ for functions $g$ and $h$ and the interaction here is the part of
the response not captured by such decomposition. For reference levels
$(a_0, b_0)$ (Section~\ref{sec:domain_definition}), we define the additive interaction contrast as
$$I(a,b;\,a_0,b_0) = \mu(a,b) - \mu(a,b_0) - \mu(a_0,b) + \mu(a_0,b_0) \in \mathbb{R}^d,$$
which vanishes for all $(a,b)$ if and only if $\mu$ is additive. In the
canonical $2\times2$ design with a $0$ baseline, individual conditions $A$ and
$B$, and their joint condition $AB$, this reduces to
$$I_{AB} = \mu_{AB} - \mu_A - \mu_B + \mu_0.$$
For continuous factors without a natural reference level, there is no scientifically meaningful choice for $(a_0, b_0)$. We therefore use functional ANOVA (Section~\ref{sec:metrics}), which avoids choosing a reference. Namely, its interaction at each condition is simply the average of the contrast across all possible reference points in the grid. The interaction is contrasted with the total effect $\mu(a,b) - \mu(a_0,b_0)$ which is what most aggregate evaluations are sensitive to.

\textbf{Interactions depend on the representation.} Scientific disciplines are often concerned with representations in a semantic space (rather than the raw output space), which then form the basis for drawing conclusions. For a fixed, deterministic readout function $g:\mathbb{R}^d \rightarrow \mathbb{R}^k$ , $I_{AB}^g = g(\mu_{AB}) - g(\mu_A) - g(\mu_B) + g(\mu_0)$. If $g$ is nonlinear, then interaction recovery in the output space does not imply recovering it after applying the readout.
Therefore, $\Vert \hat{I} - I \Vert$ may be considerably smaller than $\Vert \hat{I}^g - I ^g\Vert$, in which case evaluating model predictions only in the output space can be incomplete or misleading. 

\textbf{Relating endpoint errors to interaction errors.} Let $\hat{\mu}_c$ denote the predicted mean response under condition $c$ and $e_c=\hat{\mu}_c -\mu_c$ the endpoint error. By substituting into the definition of $I_{AB}$ we get:
$$\hat{I} - I = e_{AB} - e_A - e_B + e_0,$$
where $\hat{I}$ denotes the implied interaction based on the model predictions. These four errors have no guarantee of cancellation meaning that a small joint error does not bound the interaction error. Conversely, any errors shared across conditions may cancel out and thus small interaction errors do not indicate that individual conditional means are accurate.

Due to this, we distinguish between 'anchored' and fully generated estimates. The former relies on reference conditions while the latter relies fully on the model:
$$\hat{I}^{\text{anch}} = \hat{\mu}_{AB} - \mu_A - \mu_B + \mu_0, \space \space \hat{I}^{\text{full}} = \hat{\mu}_{AB} - \hat{\mu}_A - \hat{\mu}_B + \hat{\mu}_0.$$ The anchored error corresponds to $e_{AB}$ and isolates model error for unseen conditions, which is analogous to experimental settings in which only single factors (but not their joint) have been measured. $\hat{I}^\text{full}$ instead corresponds to using the model for the whole experiment.

\textbf{Distinguishing hallucination and other interaction failure modes.} 
\label{sec:alpha-beta}
To expose distinct failure modes, we decompose the prediction into two components: one along the true interaction and along an orthogonal component:
$$\hat{I} = \alpha I + I_\perp, \qquad \alpha = \frac{\left< \hat{I},I \right>}{\Vert I \Vert ^2}, \qquad I_\perp = \hat{I} - \alpha I, \qquad \beta = \frac{\Vert I_\perp \Vert}{\Vert I \Vert},$$
where $I_\perp$ is the part of the model prediction orthogonal to the true interaction. $\alpha$ then is the signed gain along the true interaction with $\beta$ denoting the orthogonal relative energy. Because the components are orthogonal this gives the following exact identity:
$$\mathcal{E}_I^2 = \frac{\Vert \hat{I} - I \Vert^2}{\Vert I \Vert^2}=(1-\alpha)^2 + \beta^2.$$
The zero-interaction predictor is $(\alpha, \beta) = (0,0)$ with $\mathcal{E}_I = 1.$ Thus, $\alpha$ indicates how much of the true interaction the model recovered and $\beta$ how much the model additionally hallucinated. $0<\alpha<1$ with a small $\beta$ can be interpreted as an attenuated predicted interaction,  $\alpha<0$ a reversal, $\alpha>1$ an exaggeration, and $\alpha \approx0$ with a large $\beta$ indicating error of which the structure is unrelated to the truth (Figure \ref{fig:concept}).  We note that with $m = \Vert \hat{I} \Vert / \Vert I \Vert$ and $\theta$ denoting the angle between the predicted and true interaction, we get  $\alpha = m \cos\theta$ and $\beta = m \sin\theta$. These are consequently recoverable from commonly used quantities rather than novel metrics.


\textbf{Identification and generalization regimes.} Many conditional-mean functions agree on the training support while yielding distinct interactions for held-out targets. Consequently, $I_{AB}$ is not identified by the training support alone, with inductive biases determining the prediction. Hence our experiments can be seen as evaluating this bias. 


Crucially, this distinguishes an interaction setting from the additive compositional generalization problem. Namely, with interactions, the widely studied L-shaped support for grid-based contexts \citep{wiedemer2023compositional} therefore requires function extrapolation that the observed marginals do not determine. We distinguish between generalization regimes based on what partially enables identification:
\begin{itemize}
    \item \textbf{Smoothness identified:} the held-out region of condition space is surrounded by joint observations.
    \item \textbf{Relationally identified:} factors $A$ and $B$ are each observed in other combinations.
    \item \textbf{Underdetermined:} regime in which only marginals have been observed and thereby a training support which does not provide any constraint on interactions on the held-out region.
\end{itemize}

\subsection{Metrics}
\label{sec:metrics}

We normalize errors such that a value of 1 corresponds to a trivial predictor. Specifically, we use the response scale $\text{RMS}(Y-Y_0)$ for response errors and the interaction scale $\text{RMS}(I)$ for interaction errors (Table \ref{tab:metrics}). $Y_0$ is a domain-appropriate control or reference response.

For a $2\times2$ design, $I_{AB}$ is the $AB$ response minus the additive part $\mu_A + \mu_B - \mu_0$. Our simulator uses a large 2D grid of conditions and therefore has many condition levels per factor. We
therefore obtain the same decomposition using functional ANOVA over this grid, with $Y = \tilde{Y} + I$, where $\tilde{Y}$ is the grand mean plus the two main-effect surfaces and $I$ the residual non-additive component. We also apply this to the model's generated grid $\hat{Y}$, giving $\hat{Y} = \hat{\tilde{Y}} + \hat{I}$, where $\hat{\tilde{Y}}$ is the additive part of the model's prediction and $\hat{I}$ its interaction. All grid cells are used to compute the grand and main-effect means. $\mathcal{E}_I$, $\alpha$, and $\beta$ are computed afterwards on the held-out region of condition space.

\begin{table}[t]
\centering
\caption{Metrics overview.}
\label{tab:metrics}
\footnotesize
\begin{tabular}{@{}lll@{}}
\toprule
Metric & Definition & $1$ corresponds to \\
\midrule
$\mathcal{E}_{\text{end}}$
  & $\mathrm{RMS}(\hat{Y} - Y) \,/\, \mathrm{RMS}(Y - Y_0)$
  & predicting the baseline \\
$\mathcal{E}_{\text{add}}$
  & $\mathrm{RMS}(\hat{\tilde{Y}} - \tilde{Y}) \,/\, \mathrm{RMS}(Y - Y_0)$& an error the size of the total response \\
$\mathcal{E}_I$
  & $\Vert \hat{I} - I \Vert \,/\, \Vert I \Vert$
  & predicting zero interaction \\
$\alpha, \beta$
  & Section~\ref{sec:alpha-beta}
  & -\\
\bottomrule
\end{tabular}
\end{table}

\subsection{Domain-specific interaction targets}
\label{sec:domain_definition}
 
The additive contrast is realized differently across our four studied systems (described in Section \ref{sec:datasets}): 
 
\begin{itemize}
    \item \textbf{Physiological time series:} given continuous factors without a natural reference level, we use the fANOVA decomposition directly in the waveform representation, such that $I_{AB}$ is itself a time series. 
    \item \textbf{Physiological readout:} a nonlinear physiological readout $g$
    is applied to the generated waveforms before the fANOVA decomposition, yielding $I^g_{AB}$.
    \item \textbf{Single-cell perturbation:} we use unperturbed controls as the reference, with $I_{AB}$ computed from condition-level mean expression vectors, thus yielding a vector over genes.
    \item \textbf{Molecular conformers:} as graphs with different atom counts cannot be averaged in coordinate space, we define the interaction on a readout: the mean heavy-atom radius of gyration over conformers; as reference, we use the unedited parent molecule
    \item \textbf{Drug dose--response:} the zero dose of the combined drug serves as the reference, and we define interaction relative to Bliss
    independence rather than additively (see below).
\end{itemize}

Additivity is not a sensible null for drug combinations as fractional effects are bounded to $[0,1]$, meaning simple addition of two drug effects can violate these bounds. Consequently, Bliss independence is commonly used in pharmacology, which assumes that two drugs act independently \citep{bliss1939toxicity}. Concretely, we define $E_A$ and $E_B$ as normalized (to $[0,1]$) fractional drug effects such as inhibition, where 0 indicates no effect and 1 fully inhibiting. When drugs act independently, their combined effect is 
$$E_{AB}^\text{Bliss} = E_A + E_B - E_A E_B.$$ Equivalently, Bliss independence means multiplication of surviving fractions, $1-E_{AB} = (1-E_A) (1-E_B)$, which is additive following the readout $g(E) = \log(1-E)$ and another example of interactions being representation-dependent (Section \ref{sec:setup}).
Next, we can define the interaction as the deviation from this independent effect:
$$I_{AB}^{\text{Bliss}} = E_{AB} - E^{\text{Bliss}}_{AB}.$$
Positive values then indicate drug-drug synergy seeing as inhibition is larger than expected and negative values vice versa indicate antagonism.

\subsection{Models}

We train conditional flow matching models following \cite{lipman2023flow} with minor variations across domains which we summarize in Appendix Table \ref{tab:intflow_implementation}. In the standard setting, we learn a velocity field $v_\theta (x_\tau, \tau, c)$ that transports $p_0 = \mathcal{N}(0, 1)$ to the conditional data distribution $p_1(\cdot \vert c)$ along the probability path $x_\tau = (1-\tau)z_0 + \tau z_1$ where $\tau \in [0,1]$ indicates the flow matching time. Models are trained using an MSE loss and, unless stated otherwise, sampled by numerical integration of the learned flow. Table \ref{tab:intflow_implementation} also includes conditioning, optimization, model architectures, and prediction targets. For molecular conformers, we use a published equivariant flow-matching model conditioned on molecular graphs \citep{hassan2024flow}. 

We compare three ways of obtaining a prediction for the joint condition $AB$:
\begin{itemize}

\item \textbf{Joint} model is trained and evaluated with conditioning directly on $AB$ and explicitly represents interaction.

\item \textbf{Composed} composes the learned conditional velocity fields
\begin{equation*}
v^{\text{comp}}_{AB}(x, t) = v_A(x, t) + v_B(x, t) - v_0(x, t) 
\end{equation*}
at every sampling step and is a widely-used compositional approach for unseen combinations \cite{liu2022composable,du2023reduce}. Importantly, $\hat{I}^{\text{full}}$ is not necessarily zero as the composed and single-factor fields induce different nonlinear sampling trajectories.

\item \textbf{Shared-model AB} relies on same network as used for the Composed evaluation, but instead directly queries $AB$. It thereby functions as a control which allows us to separate the effect of the shared training from the effect of the compositional step.
 \end{itemize}


\definecolor{coral}{HTML}{DC0073}
\definecolor{gridfill}{HTML}{F4F2EC}
\definecolor{gridline}{HTML}{C4C2B6}
\definecolor{axis}{HTML}{1F1F1D}
\definecolor{ablgray}{HTML}{8C8A80}
\definecolor{labelgray}{HTML}{5F5E5A}

\newcommand{\panelbx}{1.9}
\newcommand{\panelcw}{0.2375}

\newcommand{\condgrid}[6][1]{%
  \begin{scope}[shift={(#2,#3)}]
    \fill[gridfill] (0,0) rectangle (\panelbx,\panelbx);
    \ifnum#1>0
      \foreach \i in {1,...,7}{%
        \draw[gridline,line width=0.3pt] (\panelcw*\i,0) -- (\panelcw*\i,\panelbx);
        \draw[gridline,line width=0.3pt] (0,\panelcw*\i) -- (\panelbx,\panelcw*\i);}
    \fi
    \fill[pattern=north east lines,pattern color=coral]
      (\panelcw*#4,\panelcw*#5) rectangle
      (\panelcw*#4+\panelcw*#6,\panelcw*#5+\panelcw*#6);
    \draw[coral,line width=0.6pt]
      (\panelcw*#4,\panelcw*#5) rectangle
      (\panelcw*#4+\panelcw*#6,\panelcw*#5+\panelcw*#6);
    \draw[axis,line width=0.5pt] (0,0) rectangle (\panelbx,\panelbx);
  \end{scope}}

\newcommand{\chemsq}[4][1]{%
  \begin{scope}[shift={(#2,#3)}]
    \fill[gridfill] (0,0) rectangle (\panelbx,\panelbx);
    \draw[ablgray,line width=0.45pt,dash pattern=on 0.9pt off 1.3pt] (0.95,0.95) circle (0.68);
    \foreach \p in {{(0.17,0.19)},{(1.73,0.23)},{(0.19,1.71)},{(1.71,1.74)},{(0.95,0.08)}}
      {\fill[ablgray] \p circle (0.044);}
    \ifnum#1>0
      \foreach \p in {{(0.95,1.475)},{(0.425,0.95)},{(1.475,0.95)},{(0.95,0.425)}}
        {\fill[labelgray] \p circle (0.052);}
    \fi
    \coordinate (s00) at (0.605,0.605); \coordinate (s10) at (1.295,0.605);
    \coordinate (s01) at (0.605,1.295); \coordinate (s11) at (1.295,1.295);
    \ifnum#4>0 \def\PS{pobs}\def\ES{eobs}\else \def\PS{pabl}\def\ES{eabl}\fi
    \draw[\ES] (s00)--(s10); \draw[\ES] (s00)--(s01);
    \draw[hedge] (s10)--(s11); \draw[hedge] (s01)--(s11);
    \node[\PS] at (s00) {}; \node[\PS] at (s10) {}; \node[\PS] at (s01) {};
    \fill[pattern=north east lines,pattern color=coral] (s11) circle (0.10);
    \draw[coral,line width=0.8pt] (s11) circle (0.10);
    \draw[axis,line width=0.5pt] (0,0) rectangle (\panelbx,\panelbx);
  \end{scope}}

\begin{figure}[t]
\centering
\resizebox{0.8\textwidth}{!}{%
\begin{tikzpicture}[
  node distance=0pt,
  gnode/.style={circle,draw=labelgray,fill=white,line width=0.6pt,inner sep=0pt,minimum size=4.2pt},
  gedge/.style={draw=labelgray,line width=0.6pt},
  hedge/.style={draw=coral,line width=0.9pt,dash pattern=on 1.6pt off 1.2pt},
  pobs/.style={circle,draw=labelgray,fill=white,line width=0.7pt,inner sep=0pt,minimum size=5.5pt},
  pabl/.style={circle,draw=ablgray,fill=white,line width=0.7pt,inner sep=0pt,
               minimum size=5.5pt,dash pattern=on 1pt off 1pt},
  eobs/.style={draw=labelgray,line width=0.7pt},
  eabl/.style={draw=ablgray,line width=0.7pt,dash pattern=on 1.1pt off 1.1pt}]

\draw[gridline,line width=0.4pt] (7.05,0.45) -- (7.05,6.55);
\draw[gridline,line width=0.4pt] (0.30,3.47) -- (13.70,3.47);

\node[anchor=base,align=center] at (3.55,6.40) {\small Dose--response grid};
\node[anchor=base,text=labelgray] at (3.55,6.08)
  {\footnotesize drug-dose pair $\rightarrow 96D$ growth curve};
\condgrid{1.50}{4.05}{3}{3}{2}
\condgrid{3.70}{4.05}{2}{2}{3}
\node[anchor=base,text=labelgray] at (2.45,3.75) {\footnotesize $2\times2$};
\node[anchor=base,text=labelgray] at (4.65,3.75) {\footnotesize $3\times3$};

\node[anchor=base,align=center] at (10.55,6.40) {\small Perturbation graph};
\node[anchor=base,text=labelgray] at (10.55,6.08)
  {\footnotesize 1-2 genes $\rightarrow 500D$ PCA cell pop.};

\begin{scope}[shift={(8.50,4.05)}]
  \fill[gridfill] (0,0) rectangle (\panelbx,\panelbx);
  \coordinate (a1) at (0.266,1.530); \coordinate (a2) at (1.283,1.691);
  \coordinate (a3) at (1.663,0.941); \coordinate (a4) at (0.912,1.045);
  \coordinate (a5) at (0.266,0.456); \coordinate (a6) at (1.235,0.238);
  \draw[gedge] (a1)--(a4); \draw[gedge] (a2)--(a4); \draw[gedge] (a4)--(a5);
  \draw[gedge] (a3)--(a6); \draw[gedge] (a5)--(a6);
  \draw[hedge] (a2)--(a3);
  \foreach \n in {a1,a2,a3,a4,a5,a6}{\node[gnode] at (\n) {};}
  \draw[axis,line width=0.5pt] (0,0) rectangle (\panelbx,\panelbx);
\end{scope}

\begin{scope}[shift={(10.70,4.05)}]
  \fill[gridfill] (0,0) rectangle (\panelbx,\panelbx);
  \coordinate (b1) at (0.266,1.530); \coordinate (b2) at (1.283,1.691);
  \coordinate (b3) at (1.663,0.941); \coordinate (b4) at (0.912,1.045);
  \coordinate (b5) at (0.266,0.456); \coordinate (b6) at (1.235,0.238);
  \draw[gedge] (b1)--(b4); \draw[gedge] (b4)--(b5);
  \draw[gedge] (b5)--(b6); \draw[gedge] (b1)--(b5);
  \draw[hedge] (b2)--(b3);
  \foreach \n in {b1,b2,b3,b4,b5,b6}{\node[gnode] at (\n) {};}
  \draw[axis,line width=0.5pt] (0,0) rectangle (\panelbx,\panelbx);
\end{scope}

\node[anchor=base,text=labelgray] at (9.45,3.75) {\footnotesize supported};
\node[anchor=base,text=labelgray] at (11.65,3.75) {\footnotesize cold start};

\node[anchor=base,align=center] at (3.55,3.23) {\small Molecular conformers};
\node[anchor=base,text=labelgray] at (3.55,2.91)
  {\footnotesize mol.\ graph $\rightarrow N\times3$ conformer ensemble};
\chemsq{0.42}{0.88}{1}
\chemsq[0]{2.60}{0.88}{1}
\chemsq[0]{4.78}{0.88}{0}
\node[anchor=base,text=labelgray] at (1.37,0.58) {\footnotesize neighborhood};
\node[anchor=base,text=labelgray] at (3.55,0.58) {\footnotesize parents only};
\node[anchor=base,text=labelgray] at (5.73,0.58) {\footnotesize distant};

\node[anchor=base,align=center] at (10.55,3.23) {\small Physiology simulator};
\node[anchor=base,text=labelgray] at (10.55,2.91)
  {\footnotesize 2 cont.\ factors $\rightarrow 16 \times 64$ waveform};
\condgrid[0]{8.50}{0.88}{4}{4}{4}
\condgrid[0]{10.70}{0.88}{0}{0}{4}
\node[anchor=base,text=labelgray] at (9.45,0.58) {\footnotesize high--high};
\node[anchor=base,text=labelgray] at (11.65,0.58) {\footnotesize low--low};

\fill[gridfill] (4.69,0.08) rectangle (4.85,0.24);
\draw[axis,line width=0.5pt] (4.69,0.08) rectangle (4.85,0.24);
\node[anchor=base west,text=labelgray] at (4.93,0.12) {\footnotesize observed};
\fill[pattern=north east lines,pattern color=coral] (6.43,0.08) rectangle (6.59,0.24);
\draw[coral,line width=0.6pt] (6.43,0.08) rectangle (6.59,0.24);
\node[anchor=base west,text=labelgray] at (6.67,0.12) {\footnotesize held out};
\draw[ablgray,line width=0.7pt,dash pattern=on 1pt off 1pt] (8.10,0.16) circle (0.09);
\node[anchor=base west,text=labelgray] at (8.26,0.12) {\footnotesize removed};

\end{tikzpicture}
} 
\caption{Held-out designs across the four systems. Please see Section \ref{sec:datasets} for details.}
\label{fig:regimes}
\end{figure}
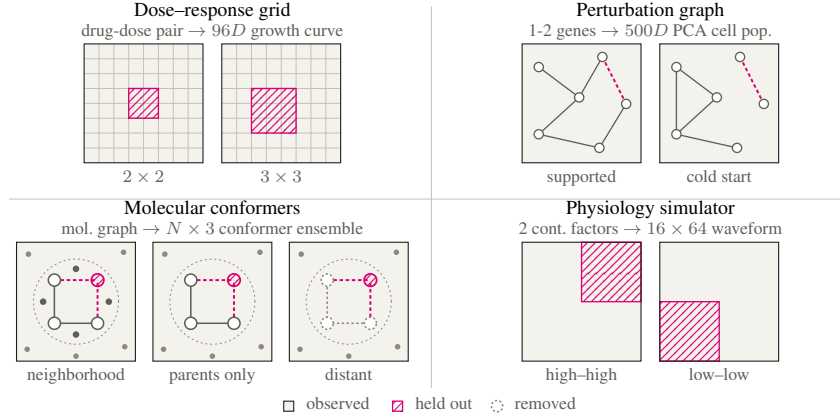

\subsection{Datasets}
\label{sec:datasets}

We study four systems, which cover different aforementioned identification regimes while having well defined interaction targets (Figure \ref{fig:regimes}). The physiological simulator of photoplethysmogram (PPG; widely used in wearables) maps melanin and blood-volume fractions to 16-channel PPG waveforms with 64 timesteps and provides exact simulator ground truths and a tunable interaction strength. Specifically, we use fANOVA to decompose the simulated grid as $Y = \tilde{Y} +I$ (Section \ref{sec:metrics}) and define $Y_\gamma = \tilde{Y} + \gamma I$. Consequently, $\gamma$ modulates the interaction while leaving both main effects and the grand mean unchanged. Then, $\gamma=1$ refers to the original grid, with $\gamma=0.5$ denoting a halved interaction strength and $\gamma=0$ a purely additive setting. We construct a simplified version of PPGen \citep{behrmann2026inferring}, where tissue composition determines optical properties in a multilayer skin model. For each channel ($ch$), we apply a standard normalized pulse-amplitude (AC/DC) readout \citep{allen2007photoplethysmography, coutrot2021perfusion}:
  \begin{equation*}
  \left(\frac{\mathrm{AC}}{\mathrm{DC}}\right)_{ch}(Y)
  =
  \frac{\max_t Y_{ch,t}-\min_t Y_{ch,t}}
       {\frac{1}{T}\sum_{t=1}^{T}Y_{ch,t}}.
  \end{equation*}
Briefly, AC is the heartbeat-synchronous change, mainly attributed to pulsatile blood volume. Meanwhile, DC is the baseline optical signal from skin, melanin, and other slowly varying effects. The readout therefore asks the physical question of how large the pulsatile optical change is relative to the baseline reflectance, and how this jointly depends on melanin and dermal blood volume.

We include three further datasets enabling investigations in measured data under smoothness-identified and relationally-identified holdouts. Specifically, the dose-response dataset consists of longitudinal yeast growth responses (96-dimensional) to pairs of drugs administered on $8\times8$ dose grids \citep{cokol2011systematic}. We hold-out and predict $2\times2$ and $3\times3$ blocks for seen drug pairs. The Norman CRISPRa Perturb-seq dataset contains single-cell responses to single- and paired gene perturbations \citep{norman2019exploring}. For the held-out double perturbations, other combinations of the targets are either included in the training set ('graph-supported' regime) or excluded ('cold-start' regime). We predict in a 500-dimensional PCA space. Finally, GEOM provides ensembles of 3D conformers for drug-like molecules \citep{axelrod2022geom}. We use chemical squares in which two edits are applied to a shared molecular core, enabling prediction of conformers for the double-edited molecule. We keep held-out targets fixed while varying the training support: parent molecules plus structurally similar molecules ('neighborhood'), 'parents only', or neither ('distant'). As edits are commonly observed on hundreds of other molecular cores, relational support is constant while local smoothness support varies.  We provide more details on these datasets in Appendix \ref{app:datasets}.  


For the empirical data, we report the anchored estimate to approximate more realistic scientific workflows. For PPG, only generated estimates are applicable. Interaction magnitudes vary across the systems, with interaction-to-total norm ratios of 5.6\% (PPG), 23.2\% (dose-response; median assay), 46.1\% (perturbation; median pair), and 47.4\% (conformer; median family). This corresponds to 0.3\%, 5\%, 21\%, and 22\% of response energy respectively. 
For $Y_0$, we use the lowest melanin/blood volume value, untreated responses, unperturbed conditions, and unedited parent molecule.

\section{Results}

\begin{figure}
    \centering
    \includegraphics[width=0.93\linewidth]{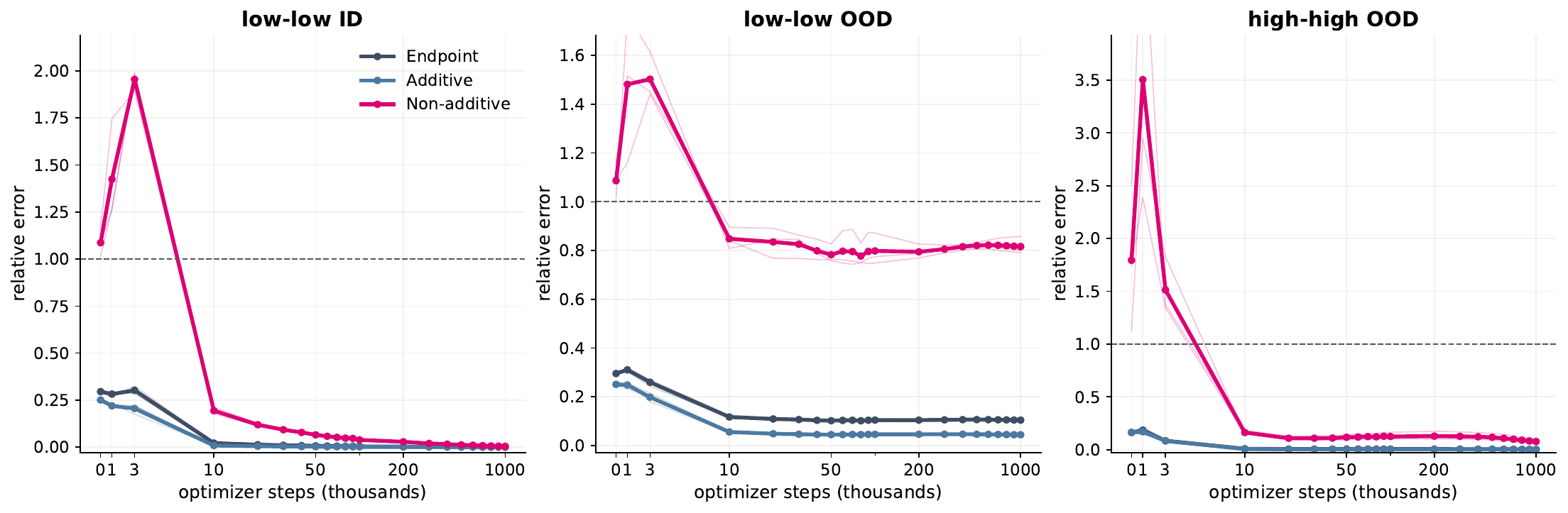}
    \caption{Learning dynamics for PPG targets in the training set (ID) and unseen targets (OOD) for separate regions of condition space. We show mean dynamics of the joint model across three random seeds, with faint lines indicating single-seed runs.}
    \label{fig:learning_dynamics}
\end{figure}

\subsection{Aggregate endpoint error conceals interaction learning dynamics}
For the synthetic PPG dataset, we hold-out two corners of the condition space (\textit{low-low} (LL) and \textit{high-high} (HH)), referring to the melanin and blood volume values) and investigate errors across training (Figure \ref{fig:learning_dynamics}). Whereas additive and endpoint errors appear to have largely converged by 10K optimizer steps, the interaction error has not. Specifically, for seen targets, the model needs at least an order of magnitude longer training. For unseen targets, we observe two distinct dynamics: when LL is held-out, interaction error stays high and plateau's by 50K steps. In contrast, for unseen HH targets, interaction error is approximately 7 times smaller by 20K steps, 

We also find such decoupling between trajectories of endpoint and interaction errors in measured data, albeit in the opposite direction. For held-out
Norman pairs, $\mathcal{E}_I$ begins to increase after ${\sim}$30K steps while
$\mathcal{E}_\text{end}$ continues to decrease until ${\sim}$50K steps
(Appendix Figure~\ref{fig:norman_learning}). Consequently, checkpoint selection using endpoint error selects models whose interaction predictions are already degrading. In the well-constrained setting of dose-response prediction, no clear decoupling is observed.

\subsection{Distinguishing interaction-prediction failures}
\begin{figure}
    \centering
    \includegraphics[width=1\linewidth]{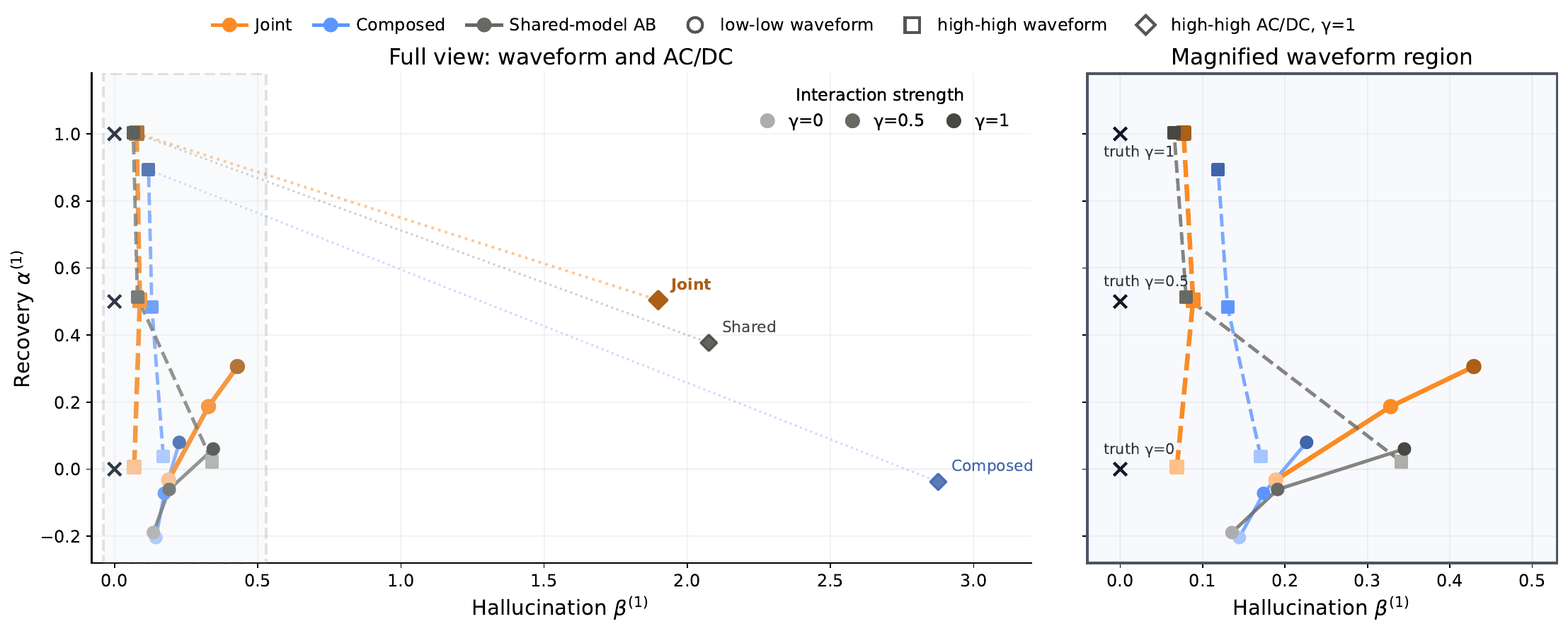}
    \caption{Interaction errors for unseen PPG targets. The left panel shows errors in the output (waveform) representation and readout (AC/DC) representation, while the right panel shows a magnified view of the (same) waveform results. The interaction strength $\gamma$ was modulated $1 \rightarrow0$ and is indicated by a $\text{dark}\rightarrow\text{bright}$ marker color gradient. Mean over training and inference seeds.}
    \label{fig:ppg_results}
\end{figure}

\textbf{Distinct recoverability of a shared condition space.} We further investigate these bifurcated results by inspecting model interaction recovery and hallucination (Figure \ref{fig:ppg_results}). We observe stark differences, with joint models largely recovering the interaction for HH but not LL targets ($\alpha\approx1.00$ and $\alpha \approx0.31$ at $\gamma=1$, respectively). Hallucinated structure is also higher for LL ($\beta \approx 0.08 \text{ vs } 0.43$). For this LL corner, we provide the produced maps of interaction in Appendix Figure \ref{fig:app_heatmaps}, showing that composed and shared-model AB baseline significantly underestimate interaction amplitude.

These results showcase that the extend to which interactions can be extrapolated based on a joint training support can strongly depend on the modeled domain. The two corners are geometrically matched but not physically equivalent. We find that a simple log-additive continuation rule selected based on the training support transfers to the held out HH corner. LL, in contrast, approaches a physical boundary where it is not (Appendix \ref{app:HH_vs_LL}). This maps to the aforementioned identifiability problem.

\textbf{Modulating interaction strengths.} The PPG simulator allowed us to do a controlled investigation of different interaction strengths ($\gamma$). However, at $\gamma=0$, we have $I=0$ and thus $\mathcal{E}_I$ and $\alpha$ are not applicable. As such, 
to quantify what interaction a model predicts in the absence of a true interaction to recover, 
we compare against the full-strength $\gamma=1$ truth $I^{(1)}$, with 
$\alpha^{(1)} = \langle \hat{I}, I^{(1)}\rangle / \Vert I^{(1)}\Vert^2$ and $\beta^{(1)} = \Vert \hat{I} - \alpha^{(1)} I^{(1)}\Vert / \Vert I^{(1)}\Vert$.
Ideally, a model shows $\alpha^{(1)} \approx \gamma$ and $\beta^{(1)} \approx 0$.


We observe that reducing interaction strength leads to monotonic decreases in $\alpha^{(1)}$, indicating models do somewhat calibrate their magnitudes based on the training support. For LL, hallucinated structure similarly decreases. Importantly, at $\gamma=0$, $\beta^{(1)}>0$ for all models. Thus, when the true interaction has been fully removed, models still generate structure, with joint $\beta^{(1)}_{LL} \approx 0.19$ and $\beta^{(1)}_{HH} \approx 0.08$. This suggests that models' inductive bias produces a baseline level of non-additivity.

\textbf{Strong predictions in output space can mask failed recovery in readout space.} While the joint model predicts interaction for unseen HH targets well in waveform space, results can differ significantly after a physiological readout, with $\alpha^{wf} \approx 1.00 \rightarrow\alpha^{AC/DC} \approx  0.50$ and $\beta^{wf} \approx 0.07 \rightarrow\beta^{AC/DC} \approx  1.90$.

\subsection{Interaction errors across regimes}
Finally, we study interaction recovery and hallucination across generalization regimes and varying training supports. For PPG, whereas previously we held out 25\% of the condition grid, we now extend this to 50\% and 75\% and train only for 100K steps due to computational costs. 

In most systems, as identification gets harder, we observe reduced interaction recovery ($\alpha$), with distinct delta magnitudes across systems. For PPG ($\gamma=1$) LL this nearly entirely removes weak recovery ($0.30 \rightarrow 0.05$), indicating the importance of locality for extrapolation. Meanwhile, HH is recoverable even far from the training support (1.03). For dose-response, we observe only a very weak drop ($0.98 \rightarrow 0.97$), with sufficient smoothness constraint persisting to the larger held-out block. We confirm this by fitting a standard (deterministic) regression baseline and find nearly identical recovery ($0.98 \rightarrow 0.97$). For perturbations, recovery is again reduced ($0.66 \rightarrow 0.41$), highlighting the importance of other observed combinations. We observe an exception for molecular conformers, with less local support leading to increases in $\alpha$ ($0.69\rightarrow0.70\rightarrow0.76$). We find this results from exaggerated differences between scaffold families, while within-family recovery remains attenuated (further details in Appendix \ref{app:geom_results}). 

Interestingly, we observe increased hallucination across systems: PPG HH $\beta$  $0.12 \rightarrow 0.25$, dose-response $0.18 \rightarrow 0.24$, perturbations $0.80 \rightarrow 0.86$, and conformers $0.63\rightarrow0.69\rightarrow0.79$. This suggests hallucinated structure not to be a fixed property of the model and that they do not merely add a baseline level of hallucinated structure, but that this scales as targets get less identifiable.

\begin{figure}[t]
    \centering

    \begin{minipage}[c]{0.55\textwidth}
        \centering
        \includegraphics[width=\linewidth]{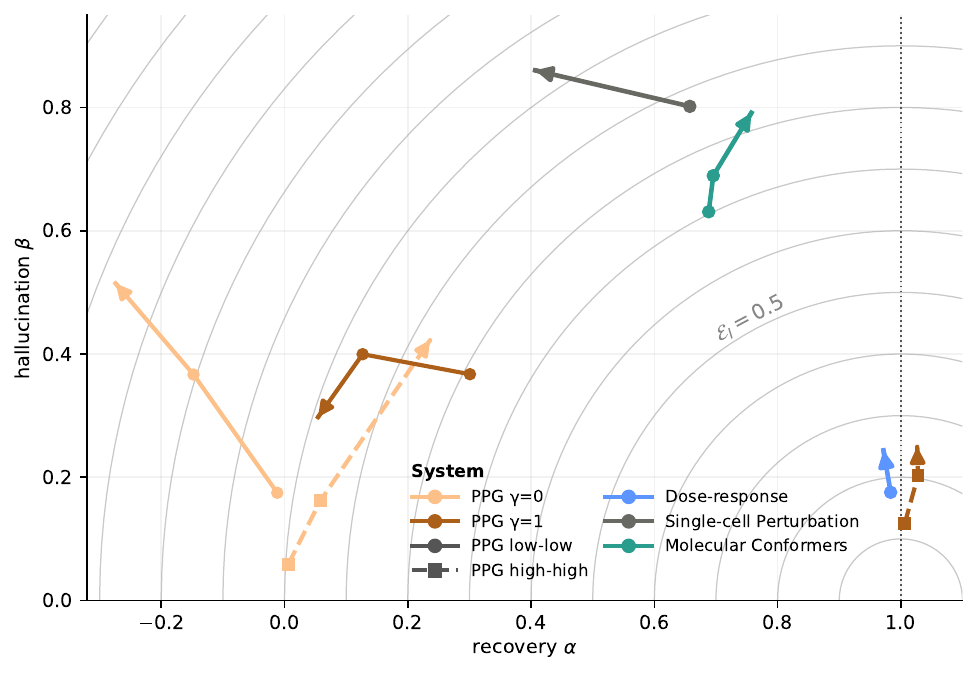}
        \caption{
            Interaction recovery and hallucination for unseen targets across systems for joint models. Contour lines indicate $\mathcal{E}_I$ and are not applicable to $\gamma=0$.
        }
        \label{fig:more_less_identified}
    \end{minipage}
    \hspace{0.001\textwidth}
    \begin{minipage}[c]{0.4\textwidth}
        \centering
        \tiny
        \begin{tabular}{lcc}
            \hline
            System / holdout & $\mathcal{E}_{\mathrm{end}}$ & $\mathcal{E}_I$ \\
            \hline
            PPG LL, 25\%   & $0.104 \pm 0.006$  & $0.798 \pm 0.065$ \\
            PPG LL, 50\%   & $0.144 \pm 0.0004$ & $0.960 \pm 0.009$ \\
            PPG LL, 75\%   & $0.165 \pm 0.007$  & $0.992 \pm 0.031$ \\
            PPG HH, 25\%   & $0.006 \pm 0.001$  & $0.125 \pm 0.035$ \\
            PPG HH, 50\%   & $0.015 \pm 0.001$  & $0.205 \pm 0.021$ \\
            PPG HH, 75\%   & $0.027 \pm 0.002$  & $0.251 \pm 0.026$ \\
            \hline
            Dose-response $2\times2$ &
                $0.045 \pm 0.002$ &
                $0.216 \pm 0.016$ \\
            Dose-response $3\times3$ &
                $0.057 \pm 0.003$ &
                $0.300 \pm 0.007$ \\
                \hline
            Perturb. supported &
                $0.582 \pm 0.027$ &
                $0.919 \pm 0.018$ \\
            Perturb. cold-start &
                $0.561 \pm 0.021$ &
                $1.052 \pm 0.044$ \\
            \hline
            Conformers neighb. & $0.436 \pm 0.003$ & $0.704 \pm 0.005$ \\
            Conformers parents & $0.467 \pm 0.015$ & $0.754 \pm 0.024$ \\
            Conformers distant & $0.513 \pm 0.003$ & $0.828 \pm 0.005$ \\
            \hline
        \end{tabular}
    \captionof{table}{Normalized endpoint and relative interaction error of joint models, with 1 indicating prediction of a control response or zero interaction respectively. Mean and SDs across training seeds. For PPG, we always use 100K step checkpoints for consistency.}
    \label{tab:e_end_e_i}
    \end{minipage}
\end{figure}

Comparing endpoint and interaction errors instead, reveals an important further insight (Table \ref{tab:e_end_e_i}). Namely, for PPG LL,  enlarging the held-out region of condition space maintains non-trivial endpoint prediction ($\mathcal{E}_\text{end}\ll1$), while making the interaction prediction no better than predicting no interaction at all ($\mathcal{E}_I\approx1$).

\section{Discussion}

We have investigated how flow models learn interactions and generate them for unseen targets using error decompositions. We have found multiple interesting dynamics: aggregate endpoint errors can conceal interaction errors, scientifically-relevant readouts can materially change interaction prediction scores, and found that both recovery and hallucination depend on identifiability regimes.

Our findings suggest that scientific evaluation should test interaction recovery directly in the representation space which is used to draw conclusions. With respect to data, individual factors appear insufficient; either collecting observations or appropriate structural assumptions are necessary to constrain unseen interactions. Computing interaction recovery and hallucination in addition to aggregate errors can help distinguish succesful extrapolation from plausible predictions which miss the scientific effects of interest.

Limitations of the presented work include our restriction to two-factor interactions and a reliance on a mean-based (rather than distributional) contrast.

\newpage

\begin{ack}

This research was funded by Gemeinnützigen Hertie-Stiftung and the Deutsche Forschungsgemeinschaft (DFG) through FOR 5187 (project number 442075332). Additional support was provided by the Machine Excellence Cluster and DFG through the Germany’s Excellence Strategy (EXC 2064 - project number 390727645) and the following projects: CRC 1404 (project number 414984028), TRR 265 (project number 402170461), and RU 5363 (project number 459422098). Finally, we are grateful to all dataset providers for access.

\end{ack}


\bibliographystyle{plainnat}
\bibliography{references}


\newpage
\appendix

\section{Interaction Maps}
\label{app:interaction_maps}

\begin{figure}[h]
    \centering
    \includegraphics[width=1\linewidth]{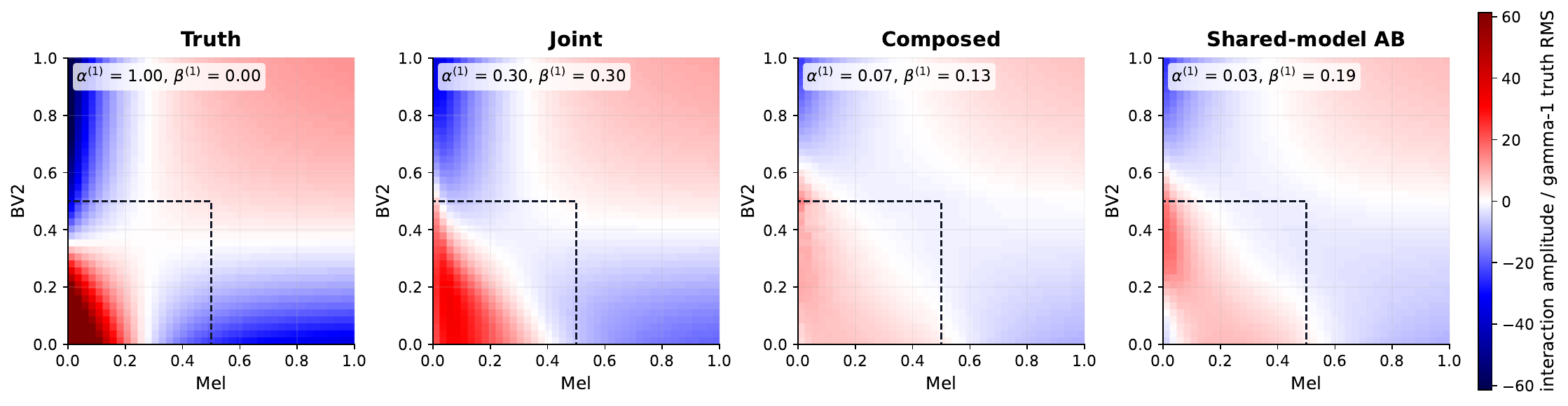}
    \caption{Maps showing the non-additive interaction projected onto the primary truth interaction mode. Colors indicate the signed interaction amplitude, normalized by the $\gamma=1$ low-low truth RMS. The dotted square indicates the held-out region of condition space (corresponding to PPG LL 25\%). To aid visualisation, we clip the colormap to [0.5\%, 99.5\%] of the maximum amplitude.}
    \label{fig:app_heatmaps}
\end{figure}

\section{Additional Learning Dynamics}
\label{app:norman_learning}

\begin{figure}[h]
    \centering
    \includegraphics[width=0.9\linewidth]{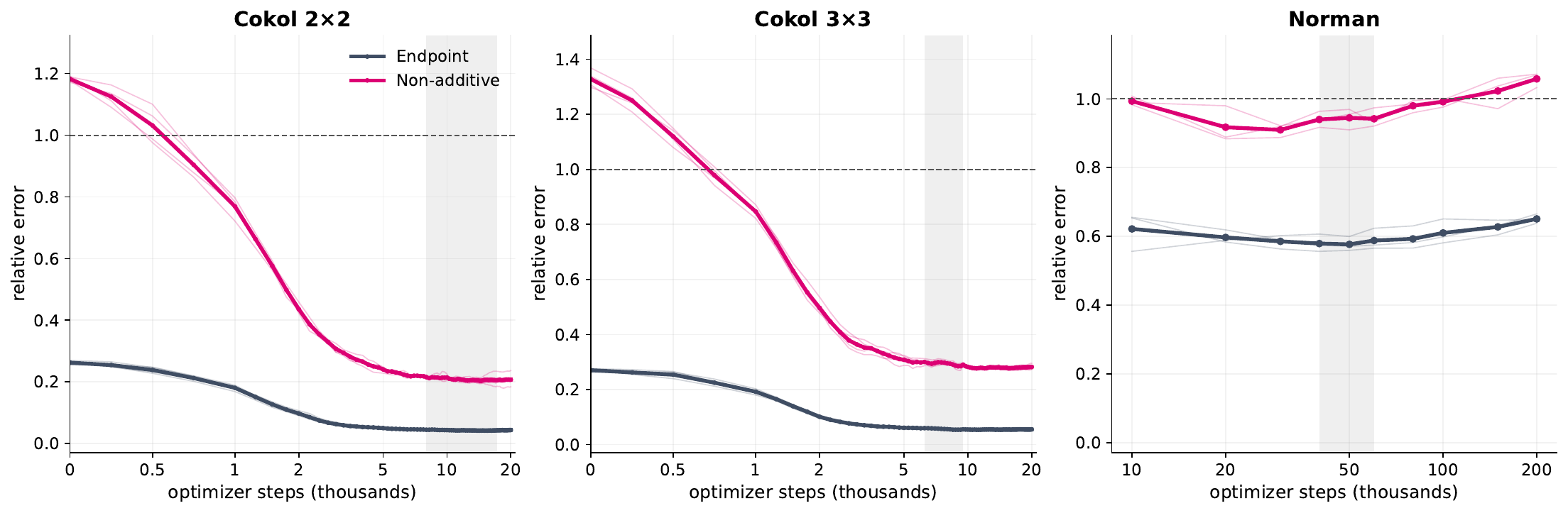}
    \caption{Held-out evaluation across the training trajectory for the \cite{cokol2011systematic} dose-response setting and for the \cite{norman2019exploring} perturbation data. While the former shows synchronous dynamics for aggregate and non-additive metrics due to a smoothness-constrained held-out region of condition space, the latter shows decoupling. Specifically, whereas $\mathcal{E}_I$ begins to increase after ~30K steps, $\mathcal{E}_{end}$ continues to decrease up to ~50K steps. The shading indicates checkpoint selection based on validation endpoint error, highlighting that using endpoint error for model selection can be suboptimal for interaction prediction.}
    \label{fig:norman_learning}
\end{figure}

\section{Dataset details}
\label{app:datasets}

\subsection{PPG simulator.} 
\label{app:ppg}

The PPG data are generated using deterministic replay of a fixed Monte Carlo photon-path bank using a three-layer skin model (simplified from \cite{behrmann2026inferring}). A path bank contains $10^8$ photon histories for 35 scattering settings with linear interpolation between adjacent settings, and Beer-Lambert reweighting applying absorption. This makes the simulator deterministic conditional on the path bank. We generate 64 time points for four source-detector distances (3-6 mm) across four wavelengths (525, 660, 850, 940 nm).

The condition space is made up of epidermal melanin ($0.25$--$14\%$) and dermal blood volume (BV2; $0.1$--$3.9\%$). Meanwhile, arterial oxygen saturation is fixed at $80\%$ and subcutaneous blood volume at $4.05\%$. The temporal blood-volume modulation follows a normalized arterial luminal-area waveform (from the PWDB database; age-45). Dermal and subcutaneous blood volumes are multiplied by $1+0.02w(t)$ with $w(t)\in[0,1]$. 

First, we simulate a uniform $101\times101$ Mel-BV2 grid. We apply balanced functional ANOVA to the clean $\gamma=1$ reference grid, writing $Y_1=\tilde{Y}+I$, where $\tilde{Y}$ contains the grand mean and the Mel and BV2 main-effect functions, and $I$ is the residual interaction. The controlled interaction-strength grids are then defined as $Y_\gamma=\tilde{Y}+\gamma I$. As such, $\gamma=1$ is the original grid, with $\gamma<1$ reducing its interaction while preserving the mean and main effects. $10^6$ Mel-BV2 pairs are drawn independently and uniformly across the condition space. We obtain the PPG waveform for every pair by bilinearly interpolating each detector-wavelength-time coordinate of the reference grid.


  \begin{table}[H]
  \centering
  \small
  \caption{PPG simulator configuration.}
  \label{tab:ppg-simulator-defaults}
  \begin{tabular}{@{}p{0.2\linewidth}p{0.66\linewidth}@{}}
  \toprule
  Component & Configuration \\
  \midrule
  Tissue geometry
    & Epidermis/dermis/subcutaneous thicknesses:
      $0.20/1.50/18.30$ mm; $0.05$-mm isotropic voxels;
      $18\times18\times20$-mm domain. \\
  Source and detectors
    & Normally incident uniform disk source, radius $1.6$ mm;
      four spherical detectors of radius $0.25$ mm at
      3, 4, 5, and 6 mm. \\
  Optical constants
    & Refractive index $n=1.4$; anisotropy $g=0.9$;
      shared reduced-scattering law
      $\mu_s'(\lambda)=6.25(\lambda/1000\,\mathrm{nm})^{-1.4}$
      cm$^{-1}$. \\
  Absorbers
    & Eumelanin in epidermis; water and lipid in all layers;
      collagen in dermis; oxygenated/deoxygenated hemoglobin
      in dermis and subcutaneous tissue. \\
  Photon-path bank
    & 35 scattering settings; $10^8$ photons per setting;
      5-ns MCX time window; fixed paths reused for absorption replay. \\
  \bottomrule
  \end{tabular}
  \end{table}



\textbf{Splitting.} We use nested low-low or high-high rectangles to create increasingly larger hold-out regions of condition space: $25\% \rightarrow 50\% \rightarrow 75\%$. These contain 735k/15k, 490k/10k, and 245k/5k training/validation samples, respectively. For final evaluation, we use a consistent held-out corner of condition space such that the test region does not change as the training support shrinks.

\subsection{Drug-response data}
\label{app:cokol}

The \cite{cokol2011systematic} yeast screenings consist of 200 drug-pair assays across 33 drugs, with 175 different-drug assays and 25 same-drug controls. Each assay is an $8\times8$ dose grid, with every cell consisting of a 96D growth curve. This results from 15 minute measurements across 24 hours, yielding 12800 total trajectories. 

\textbf{Splitting.} For each assay we hold out a fixed set of blocks (rather than cross-validation). The $2\times2$ hold-out yields [11577, 759, 464] train/validation/test trajectories, with $3\times3$ sample sizes being [11077, 769, 954]. The same 106 assays are used to compare across the two block sizes. Untreated, single-drug, and same-drug conditions are included in both training sets. 

\subsection{Single-cell perturbation data}
\label{app:norman}

The \cite{norman2019exploring} K562 CRISPRa Perturb-seq dataset consists of 91205 cells and 5045 genes, including controls, single perturbations, and 131 double perturbations which involves 73 genes. The double-perturbation conditions contain 49-1030 cells (median=257). Models use a 500D PCA representation fitted to 20000 control and single-perturbation cells. For evaluation, we decode predictions to the 5045-gene log-expression space. 

\textbf{Splitting.} We use cross-validation with four outer folds holding out each double perturbation once, while all controls and single perturbations are always included in the training set. Train/validation/test sample sizes range between [78539-80391, 2628-3770, 7348-9992] cells. Of the 131 test pairs, 113 are 'graph-supported' meaning that both genes are included in other training pairs. For the remaining 18 'cold-start' pairs, at least one of the two genes has no observed training pairs. The four folds are repeated across three training seeds.

\subsection{Molecular conformer data}
\label{app:geom}

GEOM contains conformer ensembles of drug-like molecules \citep{axelrod2022geom}. We use a pool of 26830 molecules with at least one retained conformer. The target distribution for a molecule is uniform over its 30 conformers which have the highest CREST weights (following \cite{hassan2024flow}), sampled randomly during training. As a readout, we use the heavy-atom radius of gyration ($R_g$, \AA) over conformers. For generated estimates, we average $R_g$ across 128 draws per molecule.

\textbf{Chemical squares.} Molecules are decomposes into a core with two attachment sites, each carrying one substituent. A 'square' is made up of a parent molecule (00), two single edits each swapping the substituent at one site (10, 01), and the double edit (11). From a set of 3319 squares, we retain those for which all corners share a scaffold and have $\geq 10$ conformers. We select 30 scaffold families with five squares per family, which yields 150 test queries with 384 unique parent molecules.

\textbf{Splits.} The 'neighborhood' of the test squares is specified as the union of the 30 scaffold families and all pool molecules with Morgan fingerprint (radius 2, 2048 bits, chirality-aware) Tanimoto similarity $\geq 0.6$ to a square corner (4504 molecules). Each of the three arms contain 22076 training molecules of which 17572 are shared. The \textit{Neighborhood} arm adds all 4504 neighborhood molecules (including the 384 parents), while \textit{Parents only} adds the 384 parents and 4120 additional distant controls, and finally \textit{Distant} adds only 4504 distant controls such that no parent or neighborhood molecule is seen. Controls are matched to the removed molecules across multiple graph properties (heavy atoms, rings, bonds, rotatable bonds, heteroatoms, formal charge, aromatic atoms; all absolute mean differences $\leq 0.11$). The same validation and target sets were used across the three arms.

\subsection{Evaluation}

Prior to computing endpoint and interaction metrics, we average across sampling seeds within each trained model. For PPG, we average across four sampling seeds, for drug-response data we average eight predictions, and for single-cell perturbation data we average eight batches of 24 generated cells. We report variation across training seeds, with PPG using three seeds, drug-response data uses four seeds, single-cell perturbation data using three, and conformers using three.

\section{Why do the PPG hold-out corners differ?}
\label{app:HH_vs_LL}
To understand why flow models trained on PPG data extrapolated well into the held-out high-high (HH) corner, but not into the geometrically matched low-low (LL) corner, we performed a small check on spatial continuation. Specifically, we use the training support to select from a small set of simple continuation rules. This procedure selected a raw-log-additive rule which constructed HH highly accurately (RMSE$=0.007$, interaction cosine=$1.00$). While the same rule was selected when LL was held out, it failed to extrapolate (RMSE=$0.771$). This suggests that the response dynamics in the held-out HH corner continue smoothly from observed support, while the LL region has interaction structure which cannot be accurately inferred from the L-shaped training support.

\section{Extra results for molecular conformers}
\label{app:geom_results}

\textbf{Paired differences.} As all arms predict the same queries, we can compare them via paired differences which are more precise than arm-specific means. As we move across the arms $Neighborhood \rightarrow Parents\text{ } only \rightarrow Distant$ reduces $\beta$, $\mathcal{E}_I$, adn $\mathcal{E}_\text{end}$ (family-bootstrap intervals exclude zero). Compared to $Distant$, the more supported arms also have lower $\alpha$: removing nearby support increases aligned amplitude, although this increase is outweighed by the misaligned error ($\beta$).  

\begin{table}[H]
\centering
\footnotesize
\caption{Anchored GEOM results. Arm values are mean $\pm$ SD over three training seeds. Paired differences are mean $\pm$ seed SD of per-seed differences with 95\% scaffold-family bootstrap intervals in brackets.}
\label{tab:geom_paired}
\setlength{\tabcolsep}{4pt}
\begin{tabular}{lcccc}
\toprule
 & $\alpha$ & $\beta$ & $\mathcal{E}_I$ & $\mathcal{E}_\text{end}$ \\
\midrule
Neighborhood & $0.688\pm0.014$ & $0.631\pm0.010$ & $0.704\pm0.005$ & $0.436\pm0.003$ \\
Parents only & $0.696\pm0.018$ & $0.689\pm0.018$ & $0.754\pm0.024$ & $0.467\pm0.015$ \\
Distant      & $0.758\pm0.023$ & $0.791\pm0.003$ & $0.828\pm0.005$ & $0.513\pm0.003$ \\
\midrule
Neighb.\ $-$ parents & $-0.007\pm0.018$ & $-0.059\pm0.027$ & $-0.050\pm0.029$ & $-0.031\pm0.018$ \\
 & {\scriptsize$[-0.062, 0.031]$} & {\scriptsize$[-0.122, -0.008]$} & {\scriptsize$[-0.107, -0.013]$} & {\scriptsize$[-0.066, -0.008]$} \\
Parents $-$ distant & $-0.062\pm0.025$ & $-0.102\pm0.021$ & $-0.074\pm0.025$ & $-0.046\pm0.015$ \\
 & {\scriptsize$[-0.099, -0.031]$} & {\scriptsize$[-0.181, -0.051]$} & {\scriptsize$[-0.166, -0.018]$} & {\scriptsize$[-0.103, -0.011]$} \\
Neighb.\ $-$ distant & $-0.069\pm0.009$ & $-0.161\pm0.008$ & $-0.124\pm0.007$ & $-0.077\pm0.004$ \\
 & {\scriptsize$[-0.147, -0.007]$} & {\scriptsize$[-0.281, -0.074]$} & {\scriptsize$[-0.251, -0.044]$} & {\scriptsize$[-0.155, -0.027]$} \\
\bottomrule
\end{tabular}
\end{table}

\textbf{Why does $\alpha$ increase with less support?} To understand why, we decompose $\alpha-1=[\overline{I}\,\overline{e}+\mathrm{Cov}(I,e)]/\overline{I^2}$, with endpoint error $e=\hat{\mu}_{11}-\mu_{11}$, and split the covariance into two terms: one within- and one between-family terms over the 30 scaffold families. Of the increase in $\alpha$ resulting from $Distant$ compared to $Neighborhood$ training, a global offset contributes $-0.001$, within-family terms $+0.009$, and between-family terms $+0.062$ ($\approx 90 \%$). The amplitude of family-average interactions increase from 0.87 (neighborhood) and 0.92 (parents only) overshooting to 1.14 (distant), while the within-family amplitude remains attenuated at 0.64, 0.63, and 0.65. Correlations between predicted and true interactions decrease (overall: 0.74, 0.72, 0.70, within-family: 0.70, 0.69, 0.67). This indicates that moving to a $Distant$ training setup therefore mostly rescales coarse, family-level interactions rather than boosting recovery of square-specific interactions. These results once more indicate caution is needed with single metrics and the utility of additionally inspecting $\beta$.

\section{Implementation details}
\label{app:details}

\begin{table*}[h]
\centering
\caption{Implementation details for the four experimental settings. For molecular conformers, we follow \citep{hassan2024flow}.}
\label{tab:intflow_implementation}
\scriptsize
\setlength{\tabcolsep}{4pt}
\renewcommand{\arraystretch}{1.12}
\begin{tabularx}{\textwidth}{@{}
  >{\raggedright\arraybackslash}p{0.14\textwidth}
  >{\raggedright\arraybackslash}X
  >{\raggedright\arraybackslash}X
  >{\raggedright\arraybackslash}X
  >{\raggedright\arraybackslash}X@{}}
\toprule
\textbf{Detail} & \textbf{PPG} & \textbf{Dose-response} & \textbf{Perturbation} & \textbf{Conformers} \\
\midrule
Output & $16\times64$ waveform & 96-point trajectory & PCA-500 expression & 3D coordinates, all atoms \\
Architecture & Conv-patch DiT & Symmetric MLP & Residual MLP & ET-Flow equivariant transformer (O(3), post-hoc parity) \\
Width / depth & 128 / 4 & 256 / 3 & 512 / 4 & 160 / 20 \\
Attention heads & 4 & N/A & N/A & 8 \\
Parameters & 0.86M (both) & 0.51M each & 3.03M & 8.34M \\
Conditioning & 2 factors; + presence bits (shared) & Drug-token set & Gene-token set & Molecular graph (atom/bond features) \\
Condition set & $AB$ (Joint); all four (shared) & 0--2 drugs & 0--2 genes & One graph per molecule \\
Flow source & Gaussian noise & Gaussian noise & Measured control (OT-paired) & Harmonic prior, $\sigma=0.1$, RMSD-aligned \\
Prediction target & Endpoint $x_1$ & Endpoint $x_1$ & Velocity $x_1-x_0$ & Velocity \\
Training loss & MSE; 4-mask MSE (shared) & Endpoint MSE & Velocity MSE & Velocity MSE \\
Optimizer & AdamW & AdamW & AdamW & AdamW \\
Learning rate & $2\times10^{-4}$ & $2\times10^{-4}$ & $2\times10^{-4}$ & $5\times10^{-4}$ (peak) \\
Weight decay & $10^{-8}$ & $10^{-4}$ & $10^{-4}$ & $10^{-8}$ \\
LR schedule & 5k warm-up + cosine & Constant & Constant & 1k warm-up + cosine to $10^{-6}$ \\
Gradient clipping & None & 5 & 1 & Adaptive (ET-Flow gradient-norm queue) \\
EMA decay & 0.999 & 0.999 & 0.999 & None \\
Batch size & 512 & 512 & 512 & 32 molecules ($16\times2$ accumulation) \\
Training budget & 1M steps & 20k steps & 200k steps / fold & 200 epochs (138k steps) \\
Checkpoint & Final EMA & Best covered-validation EMA & Best inner-validation EMA & Final (epoch 200) \\
Primary inference & Euler & $t=0$ endpoint & Heun & Euler \\
Inference steps & 128 & 1 & 20 & 50 \\
Joint prediction & Direct AB query & Direct AB query & Direct AB query & Direct query of the 11 graph \\
Composed prediction & $\hat{x}_A+\hat{x}_B-\hat{x}_{\emptyset}$ & N/A & $v_A+v_B-v_{\emptyset}$ & N/A \\
\bottomrule
\end{tabularx}
\end{table*}

\end{document}